\documentclass[10pt,twocolumn,letterpaper]{article}

\usepackage{cvpr}              

\usepackage[accsupp]{axessibility}

\usepackage{indentfirst}

\usepackage[top=2cm, bottom=2cm, left=2cm, right=2cm]{geometry}
\usepackage{algorithm}
\usepackage{algorithmicx}
\usepackage{algpseudocode}
\usepackage{amsmath}

\usepackage{multirow}
\usepackage{graphicx}
\usepackage[normalem]{ulem}
\usepackage{diagbox}
\useunder{\uline}{\ul}{}

\usepackage{caption}

\usepackage[dvipsnames]{xcolor}

\definecolor{cvprblue}{rgb}{0.21,0.49,0.74}
\usepackage[pagebackref,breaklinks,colorlinks,citecolor=cvprblue]{hyperref}

\def\paperID{9257} 
\def\confName{CVPR}
\def\confYear{2024}

\title{PAD: Patch-Agnostic Defense against Adversarial Patch Attacks}
\author{Lihua Jing\textsuperscript{1,2}, Rui Wang\textsuperscript{1,2}\thanks{Corresponding author} , Wenqi Ren\textsuperscript{3}, Xin Dong\textsuperscript{1,2}, Cong Zou\textsuperscript{1,2}\\
\textsuperscript{1}Institute of Information Engineering, Chinese Academy of Sciences\\
\textsuperscript{2}School of Cyber Security, University of Chinese Academy of Sciences\\
\textsuperscript{3}School of Cyber Science and Technology, Shenzhen Campus of Sun Yat-sen University\\
{\tt\small \{jinglihua, wangrui, dongxin, zoucong\}@iie.ac.cn, renwq3@mail.sysu.edu.cn}
}

\begin{document}
\maketitle
\begin{abstract}
Adversarial patch attacks present a significant threat to real-world object detectors due to their practical feasibility.
Existing defense methods, which rely on attack data or prior knowledge, struggle to effectively address a wide range of adversarial patches.
In this paper, we show two inherent characteristics of adversarial patches, semantic independence and spatial heterogeneity, independent of their appearance, shape, size, quantity, and location. Semantic independence indicates that adversarial patches operate autonomously within their semantic context, while spatial heterogeneity manifests as distinct image quality of the patch area that differs from original clean image due to the independent generation process.
Based on these observations, we propose PAD, a novel adversarial patch localization and removal method that does not require prior knowledge or additional training.
PAD offers patch-agnostic defense against various adversarial patches, compatible with any pre-trained object detectors. 
Our comprehensive digital and physical experiments involving diverse patch types, such as localized noise, printable, and naturalistic patches, exhibit notable improvements over state-of-the-art works.
Our code is available at \href{https://github.com/Lihua-Jing/PAD}{https://github.com/Lihua-Jing/PAD}.

\end{abstract}    
\vspace{-1.5mm}
\section{Introduction}
\label{sec:intro}

Adversarial attacks substantially challenge the security of object detectors, leading to potentially severe consequences in various fields (\textit{e.g.}, autonomous driving). Traditional adversarial attacks typically involve adding perturbations to the entire image. However, modifying every pixel is unrealistic in real-world attack scenarios. Adversarial patch attacks, on the other hand, focus on introducing disturbances in a limited area. Their practical feasibility makes them one of the most threatening forms of adversarial attacks.

\begin{figure}[t]
    \centering
    \includegraphics[width=0.48\textwidth]{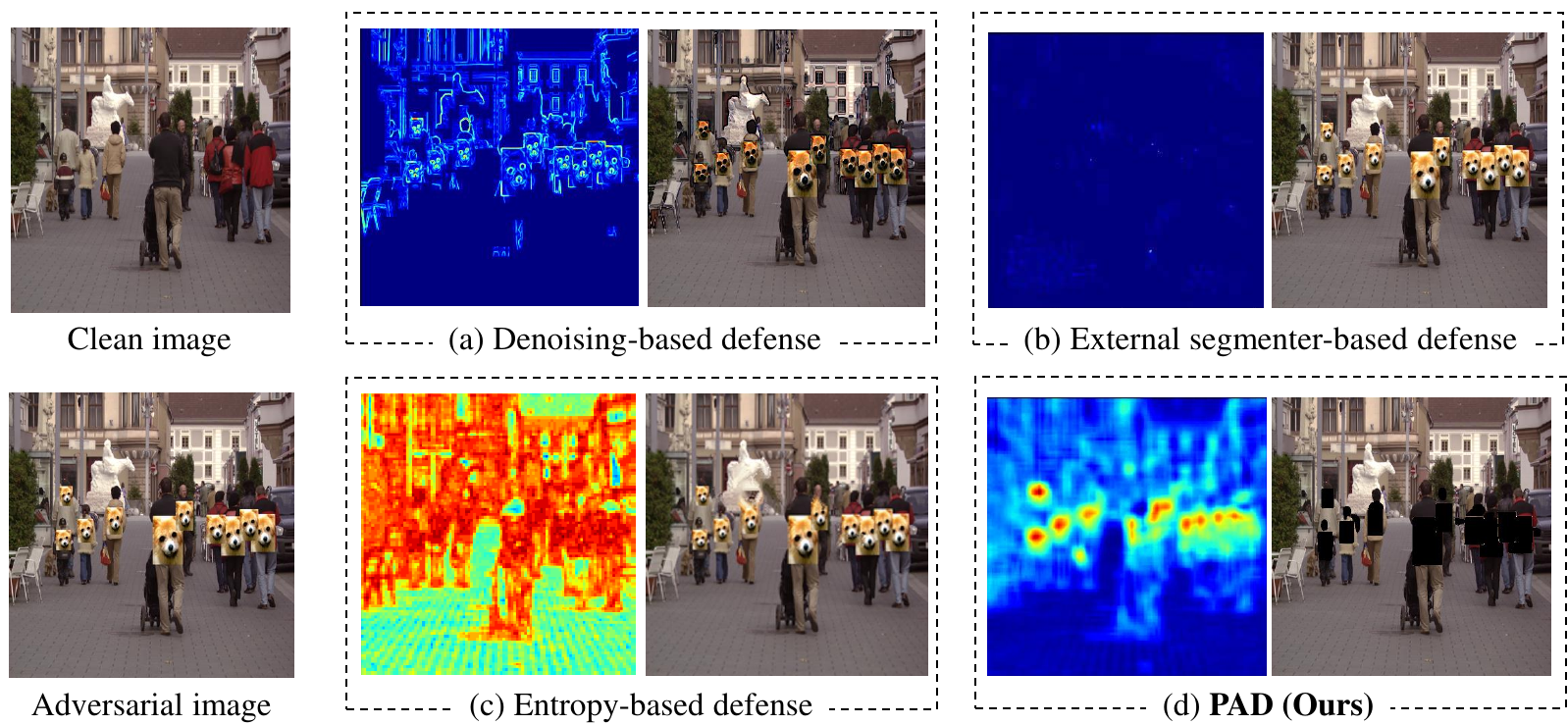}
    \caption{When attacked by natural-looking adversarial patches, (a) locates high-frequency areas, and eliminates edge lines instead of patches \cite{naseer2019local}; (b) fails to detect the existence of patches since no such patch data in the training set \cite{liu2022segment}; (c) produces a heat map where the patch area and background are difficult to distinguish, failing to locate the patches \cite{tarchoun2023jedi}. Our proposed PAD achieves accurate patch location and removal.}
    \vspace{-4mm}
    \label{fig:intro}
\end{figure}

Defenses against adversarial patch attacks on object detectors can be broadly categorized into three main types: \textit{i)} modifying or intervening within detection models \cite{saha2020role,ji2021adversarial,kim2022defending}, \textit{ii)} locating and eliminating adversarial patch regions in images\cite{naseer2019local,chiang2021adversarial,liu2022segment,tarchoun2023jedi,bunzel2023adversarial}, and \textit{iii)} certiﬁably robust defenses \cite{xiang2021detectorguard,xiang2023objectseeker}. 
Among these,  methods falling under the second category, which act as preprocessing, offer the broadest range of applications. 

Researchers have explored various patch localization methods to effectively remove adversarial patches from images. Denoising-based defenses \cite{naseer2019local} smooth out noise-like regions in images, providing an effective defense against early localized noise patches. However, they fail to address natural-looking patches. 
External segmenter-based defenses \cite{chiang2021adversarial,liu2022segment} train an adversarial patch segmentation model for patch localization, using adversarial images generated by existing attack techniques. However, the reliance on training data makes them ineffective against unseen patch types. 
Entropy-based defenses \cite{tarchoun2023jedi,bunzel2023adversarial} achieve patch localization by identifying high entropy kernels and patch shape reconstruction. Nevertheless, the entropy threshold setting requires prior knowledge of the distribution for clean data and patches, and shape reconstruction relies on training data, posing challenges in practical applications. 
Despite progress in certain aspects, these methods face a common challenge of locating various adversarial patches without relying on prior attack knowledge. As shown in Figure \ref{fig:intro}, these three categories of methods fail to effectively remove patch areas.

In this paper, we propose a new approach for various adversarial patch localization without relying on prior attack knowledge (\textit{e.g.,} appearance, shape, size, or quantity). The proposed approach is derived from two inherent characteristics of adversarial patches: semantic independence and spatial heterogeneity.

Semantic independence implies zero information gain from the surrounding semantic space, while spatial heterogeneity refers to inconsistent image quality in the same space introduced by adversarial patches.
In both digital and physical attacks, the adversarial patch, added as a separate component to the image or environment, is semantically independent in the image. The surroundings provide no information about the content of adversarial patches, and vice versa. Additionally, different imaging devices, generation processes, and compression methods may lead to variations in image quality. With a source different from the original clean data, the quality of patch regions exhibits heterogeneity compared to other areas in space.

Based on these observations, we propose to identify patch areas by quantifying local semantic independence and spatial heterogeneity. We measure the information gain between adjacent regions based on mutual information, and evaluate residuals from recompression at different quality factors to address the unequal impact upon areas of varying quality. While complex backgrounds confuse entropy-based methods \cite{tarchoun2023jedi,bunzel2023adversarial} since the high information density caused by complicated textures, our method exhibits more robustness as semantic correlations remain between adjacent background areas.
In addition, to eliminate the reliance on training data, we present a patch localization and removal pipeline that requires no prior knowledge or additional training. Different from current works, our defense accurately identifies the patch region mask without any reference to existing adversarial images and imposes no limitations on the quantity or proportion of patches in the image.

Our contributions can be summarized as follows:

\begin{itemize}
    \item We reveal two inherent characteristics of adversarial patches, semantic independence and spatial heterogeneity, and propose patch locating based on mutual information and recompression, which is agnostic to patch appearance, shape, size, location, and quantity.
    \item We propose a patch-agnostic defense (PAD) method for adversarial patch localization and removal, which requires no prior attack knowledge or additional training and is compatible with any object detector.
    \item We conduct experiments on adversarial patches with different appearances, shapes, sizes, locations, and quantities, evaluating the defense effectiveness of PAD in both digital and physical scenes. Experimental results demonstrate our superior defense performance compared to the current state-of-the-art methods.
\end{itemize}

\section{Related Work}
\label{sec:formatting}

\begin{figure*}[t]
    \centering
    \includegraphics[width=1\textwidth]{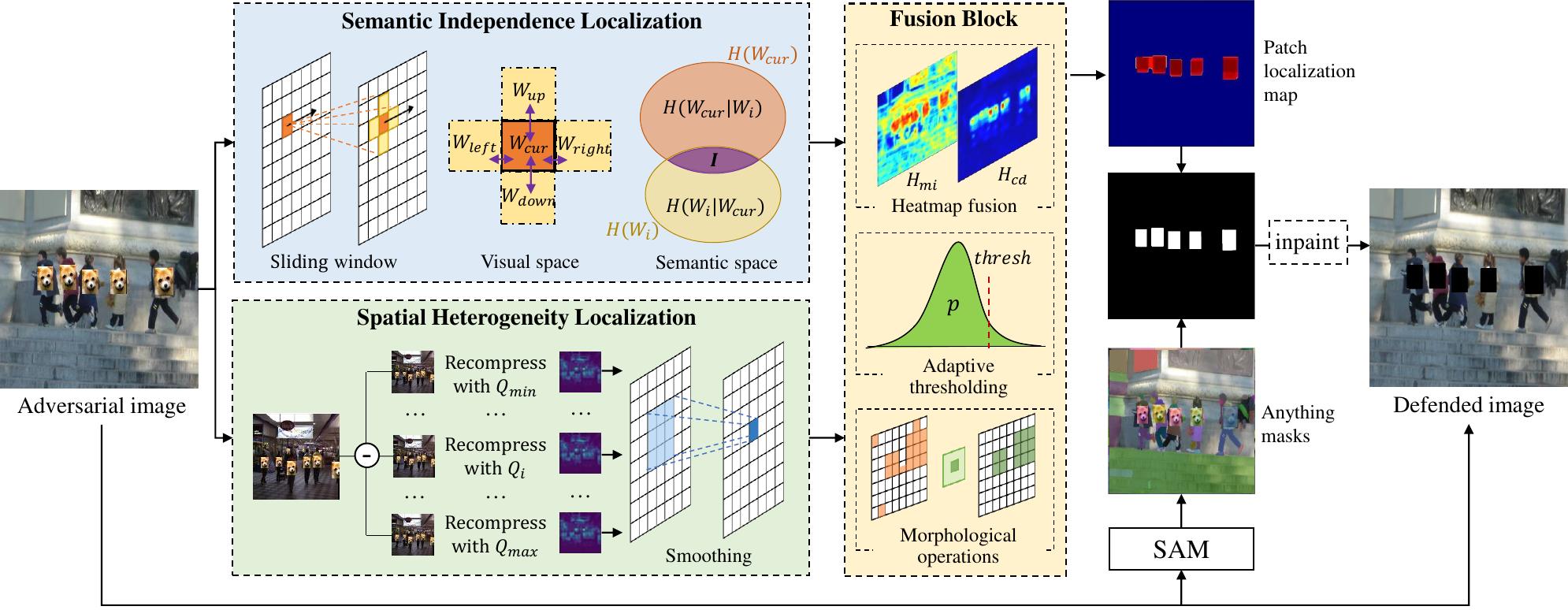}
    \caption{Overview of our proposed PAD. Semantic Independence Localization and Spatial Heterogeneity Localization find patch regions from two views, generating heat maps feeding into Fusion Block. The patch localization map output by Fusion Block is then matched with all masks from SAM, getting more accurate patch boundaries. Feeding the defended image into object detectors for robust prediction.}
    \vspace{-2mm}
    \label{fig:method}
\end{figure*}

\subsection{Adversarial Patch Attacks}

The concept of adversarial patch attacks is first introduced by \cite{brown2017adversarial}. They develop a generic patch capable of deceiving image classifiers and demonstrate the feasibility of physical attacks by attaching the patch in real-world scenarios. 
Early research on adversarial patch attacks primarily focuses on localized noise \cite{karmon2018lavan,liu2018dpatch}. 
DPatch \cite{liu2018dpatch} is the pioneering work on patch attacks specifically designed for object detection, targeting both bounding box regression and object classification components of the detection system. PatchAttack \cite{yang2020patchattack} proposes a reinforcement learning-based attack method to induce misclassification by superimposing small texture patches on the input image. 
Some work focus on physical attacks \cite{zhao2019seeing,chen2019shapeshifter,thys2019fooling,xu2020adversarial,wu2020making}.  \cite{zhao2019seeing} and \cite{chen2019shapeshifter} attach patches to traffic signs, leading to the misidentification of those signs. \cite{thys2019fooling} proposes a printable adversarial patch for pedestrian detection, introducing non-printable loss in the optimization process. \cite{xu2020adversarial} and \cite{ wu2020making} explore the integration of adversarial patches into wearable clothing.
\cite{hu2021naturalistic}, \cite{kong2020physgan} and \cite{tan2021legitimate} train generative adversarial networks (GANs) to generate natural-looking patches that match the visual properties of normal images.

\subsection{Defenses against Patch Attacks}

Adversarial training \cite{jakubovitz2018improving,shafahi2019adversarial,wu2019defending,rao2020adversarial,zhang2019towards,gittings2020vax,metzen2021meta}, which enhances model robustness by adding adversarial examples during training, is one of the most popular and effective defenses against digital attacks. However, such methods are not suitable for defending pre-trained models already in use and require significant resources for retraining when new attacks emerge, making them not so practical.

Some defense methods involve modification of specific models \cite{saha2020role,ji2021adversarial,yu2021defending,kim2022defending}. \cite{saha2020role} investigates the use of spatial context constraints in YOLOv2 \cite{redmon2017yolo9000} to enhance defense robustness against adversarial patches. \cite{ji2021adversarial} introduces a patch class into YOLOv2, enabling the detection of objects of interest as well as adversarial patches. \cite{kim2022defending} proposes adversarial patch feature energy(APE), and defense is achieved by incorporating an APE discovery and suppression module into the network. Although good defense effects can be achieved on specific detection models, they cannot directly provide defense for a wide range of object detectors.

To provide more general defense, researchers have explored locating patch areas in images and eliminating their effects \cite{rossolini2023defending,naseer2019local,chiang2021adversarial,liu2022segment,hayes2018visible,chen2021turning,tarchoun2023jedi,bunzel2023adversarial,xu2023patchzero}. Since early adversarial patches are usually in the form of localized noise, some defense methods focus on reducing the impact of noise-like areas in input images. LGS \cite{naseer2019local} observes that patch attacks introduce concentrated high-frequency noise and proposes gradient smoothing for regions with significant gradients. APM \cite{chiang2021adversarial} and SAC \cite{liu2022segment} train external segmentation networks to locate noise-like regions. While these methods effectively defend against localized noise-based patch attacks, they struggle to counter the new types of natural-looking patches. DW \cite{hayes2018visible} and Jujutsu \cite{chen2021turning} utilize saliency maps to identify patch areas and cover them to mitigate their impact on classification. In object detection tasks, which involve bounding box regression in addition to classification, accurately localizing patches becomes challenging using saliency maps. Jedi \cite{tarchoun2023jedi} and \cite{bunzel2023adversarial} use entropy to locate patch areas, but prior knowledge of entropy distribution of clean dataset and patch area is required.

In recent years, some researchers have proposed certifiably robust defenses against adversarial patches \cite{xiang2021detectorguard,xiang2023objectseeker,xiang2021patchguard,xiang2021patchguard++,xiang2022patchcleanser}. DetectorGuard \cite{xiang2021detectorguard} is an attack detection defense that raises an alert when an attack is detected without removing the adversarial patches, resulting in a loss of model functionality during an attack. ObjectSeeker \cite{xiang2023objectseeker} requires ensuring that, under at least one partition, the remaining images do not contain any adversarial patch pixels. As a result, there will be trouble handling attack scenarios with large patch proportions or multiple patches.

Different from these methods, PAD is derived from two general characteristics of patches that are independent of their appearance, shape, size, quantity, and location, allowing us to eliminate various patches without relying on prior attack knowledge.

\section{Preliminaries}
\label{sec:formatting}


Differing from traditional adversarial attacks, adversarial patch attacks impose restrictions on the attacker, limiting the area where perturbations can be introduced. Within this constraint, the attacker has the flexibility to manipulate the pixels within the designated patch region.

We denote a clean image with dimensions $w\times h\times c$ as $X\in \mathbb{R} ^{w\times h\times c}$. The adversarial patch, denoted as $P_{adv}$, can take any shape. The generation of $P_{adv}$ is typically controlled by the loss function $L_{patch}$, which varies depending on the specific attack objective. Since our goal is to defend against various patch attacks, irrespective of their intention to conceal the target object, misclassify it, or generate false detections of non-existent objects, we do not make assumptions about $L_{patch}$.

The resulting adversarial image $X_{adv} \in \mathbb{R} ^{w\times h\times c}$, produced by attack techniques, can be expressed as follows:
\begin{equation}
X_{adv} = M_{patch}\odot A(P_{adv},X,l,t) + (1-M_{patch})\odot X,
\end{equation}
where $M_{patch}\in\left\{0,1\right\}^{w\times h}$ represents the patch region in image $X$, with elements set to 1 within $P_{adv}$ and 0 elsewhere. $A(P_{adv},X,l,t)$ denotes the patch application function, incorporating patch transformations such as scaling and rotation denoted by $t$, and patch location denoted by $l$. $\odot$ refers to element-wise multiplication. In the case of attack methods that can be used physically, $A(P_{adv},X,l,t)$ typically involves completely replacing the image area at position $l$ with the transformed patch.


\section{Method}
\label{sec:intro}

\subsection{Defense Pipeline}

In this section, we introduce the pipeline of PAD, as shown in Figure \ref{fig:method}. Firstly, we analyze the input image using the two inherent characteristics that all adversarial patches possess to obtain heat maps, $H_{mi}$ and $H_{cd}$, which highlight the regions in the image that exhibit semantic independence and spatial heterogeneity, respectively. Next, we employ the Fusion block to merge $H_{mi}$ and $H_{cd}$, generating the patch localization map $H_{p}$ that accurately reflects the areas in the image that possess both of these characteristics.

To make the patch masks more accurate, we introduce the Segment Anything Model (SAM) \cite{kirillov2023segment}. Different from the segmentation models introduced in prior methods, since we do not need it to have recognition capabilities for adversarial patches, SAM can be replaced with any pre-trained segmentation models with similar capabilities, without additional training. In other words, we do not rely on known patch attack methods to generate adversarial images for training, preventing our defense from losing effectiveness when encountering new attack methods. With SAM's zero-shot segmentation capability, we segment the edges of all regions in the image and obtain masks for each region. We then match each mask with $H_{p}$ and consider all masks with Intersection over Area (IoA) greater than threshold $t_{m}$ as the final patch masks.
The calculation of IoA can be stated as follows:
\begin{equation}
IoA(mask, H_{p})=\frac{area(mask\bigcap H_{p})}{area(mask)}. 
\end{equation}

If the localization of adversarial patches is accurate enough, the removal process only needs to be able to eliminate the impact of the patches. Therefore, we employ a simple and fast inpainting method that is commonly used in previous works \cite{liu2022segment,chiang2021adversarial}: filling the patch area with all black pixels. We also compare the coherence transport-based inpainting method \cite{bornemann2007fast,tarchoun2023jedi} with all-black, more details can be found in supplementary material.

\subsection{Semantic Independence Localization}

\label{subsec:semantic_independence}

\noindent\textbf{Semantic independence evaluation.} The semantic independence of a region can be measured by its semantic correlation with surrounding regions. The smaller the semantic correlation, the stronger the independence. For two adjacent regions, $A$ and $B$, their semantic correlation can be defined as the information gain they provide to each other. Given knowledge about region $A$, it quantifies how much information can be inferred about region $B$, i.e., the reduction in uncertainty about $B$ given knowledge of $A$. 
This can be expressed using mutual information:
\begin{equation}
I(A;B)=H(A)-H(A|B)=H(B)-H(B|A)
\end{equation}
where $H(\ast)$ represents information entropy, and $H(\ast|\ast)$ represents conditional entropy.

\noindent\textbf{Semantic independence localization based on mutual information.}  Building on the analysis above, we perform adversarial patch discovery by computing the semantic independence of local regions across the entire image. We set up a sliding window, and calculate the mutual information between each window and its four neighboring windows (up, down, left, right). The average value of these mutual information scores is used as the heat value for the current window, generating a heat map for the entire image. We use $W_{cur}$ to denote the current window, and  $W_{up}$,  $W_{down}$,  $W_{left}$,  $W_{right}$ represent the four neighboring windows of the same size respectively. For each $i$ in $\left \{up, down, left, right  \right \}$, the mutual information between $W_{i}$ and $W_{cur}$ can be expressed as follows:
\begin{equation}
I(W_{i};W_{cur})=\sum_{w_{i}\in W_{i}}\sum_{w_{c}\in W_{cur}}p(w_{i},w_{c})\log_{}{\frac{p(w_{i},w_{c})}{p(w_{i})p(w_{c})} }. 
\end{equation}

The heat value within $W_{cur}$ can be expressed as follows:
\begin{equation}
H_{mi}\left [ x_{cur}: x_{cur}+d,  y_{cur}: y_{cur}+d\right ]  =\frac{1}{n} {\sum_{i=1}^{n}I(W_{i};W_{cur}) },
\end{equation}
where$(x_{cur}, y_{cur})$ denotes the coordinates of the upper left corner of the window $W_{cur}$, $H_{mi}$ denotes the heat map generated by Semantic Independence Localization module, $d$ denotes the size of the sliding window, and $n$ denotes the number of neighboring windows. $n$ equals 4 for most windows, 2 or 3 for windows located at the edge of the image.

\subsection{Spatial Heterogeneity Localization}

\label{subsec:spatial_heterogeneity}


\noindent\textbf{Impact of compression on image quality.} Image compression leverages the insensitivity to certain components of human eyes to reduce storage space. Taking the most commonly used JPEG compression as an example, after color space transformation, it undergoes block-based Discrete Cosine Transform (DCT) to the image, converting the spatial domain into the frequency domain. The transformed low-frequency components have larger values, mainly concentrated in the upper-left corner, while high-frequency components have smaller values, distributed in lower-right regions. Subsequently, the DCT coefficients are quantized so that smaller coefficients close to 0 completely become 0, and non-zero values also generate a large number of repetitions, thereby reducing the coding length. Using $F(x,y,i)$ to represent the DCT coefficient of channel $i$ at location $(x,y)$, the quantization process can be expressed as:
\vspace{-0.2mm}
\begin{equation}
F_{q}(x,y,i) =round(\frac{F(x,y,i)}{Q(x,y,i)} ) ,
\end{equation}
where $Q(x,y,i)$ represents the corresponding quantization step size. A larger $Q$ leads to greater quantization loss and poorer image quality.

\noindent\textbf{Spatial heterogeneity localization through recompression.} For an image containing regions of varying quality, compressing the entire image will affect each quality region differently, providing valuable clues for identifying abnormal regions in the image \cite{lukavs2003estimation,lin2011passive,bianchi2012image,farid2009exposing,he2006detecting}. Inspired by this, we locate adversarial patches based on the quality differences during recompression. For a clean image with quality factor $Q_{c}$, the patch area with quality factor $Q_{p}$, we set different quality factors  $Q_{r}$ to re-compress the attacked image $X_{adv}$, and calculate the squared difference of pixel values before and after re-compression, as follows:
\vspace{-0.2mm}
\begin{equation}
D(x,y,Q_{r})=\frac{1}{c} \sum_{i=1}^{c} \left [ f(x,y,i)- f_{Q_{r}} (x,y,i) \right ] ^{2} ,
\end{equation}
where $c$ denotes the number of channels.
When $Q_{r}$ is close to $Q_{p}$, the $D$ values of the patch region are minimized. When $Q_{r}$ is close to $Q_{c}$, the $D$ values of the uncovered region are minimized. 
To enhance the robustness to texture variations in the image, we apply convolutional smoothing and perform normalization.

\subsection{Fusion Block}

Section \ref{subsec:semantic_independence} and \ref{subsec:spatial_heterogeneity} aim to identify potential adversarial patch regions from the perspectives of semantic independence and spatial heterogeneity. In the fusion block, we merge the heat maps obtained from these two methods. Additionally, to mitigate the influence of cluttered backgrounds, we apply adaptive thresholding and morphological operations to further process the fused results.

\noindent\textbf{Heatmap fusion.} Given the local mutual information heat map $H_{mi}$ and the recompression difference heat map $H_{cd}$ which have different value ranges, we first normalize them individually to scale the values into the range $\left [ 0,255 \right ] $. The normalization process can be expressed as follows:
\begin{equation}
H_{mi}^{'} (x,y)=\frac{H_{mi}(x,y)-min(H_{mi})}{max(H_{mi})-min(H_{mi})} \times 255,
\end{equation}
\begin{equation}
H_{cd}^{'} (x,y)=\frac{H_{cd}(x,y)-min(H_{cd})}{max(H_{cd})-min(H_{cd})} \times 255.
\end{equation}
After normalization, we perform a weighted sum of $H_{mi}^{'}$ and $H_{cd}^{'}$ to obtain the fused heat map, $H_{fuse}^{'}$:
\begin{equation}
H_{fuse} (x,y)=r_{mi}\times  H_{mi}^{'} (x,y)+(1-r_{mi})\times H_{cd}^{'} (x,y),
\end{equation}
where $r_{mi}\in \left [ 0,1 \right ] $ denotes the weight of  mutual information heat map.

\noindent\textbf{Adaptive thresholding.} Since the heat values are significantly affected by the image content, using a static threshold may filter out adversarial patches or incorrectly treat other regions as adversarial patches, thus degrading the performance of the model. Therefore, we automatically set an adaptive threshold based on the distribution of the heat map for each image, and then set elements in $H_{fuse}$ that are below the threshold to 0. The threshold here can be expressed as follows:
\begin{equation}
thresh=(1-j)\times Sort(H_{fuse})[i]+j\times Sort(H_{fuse})[i+1],
\end{equation}
\begin{equation}
i=\left \lfloor (n-1)\times p \right \rfloor , j=(n-1)\times p-i,
\end{equation}
where p is a fixed hyperparameter, n represents the number of elements in $H_{fuse}$, and $Sort(H_{fuse})$ represents sorted $H_{fuse}$ in ascending order based on the heat values.

\noindent\textbf{Morphological operations.} To eliminate the interference of background with high heat values but unrelated to adversarial patches, we apply an OPEN-CLOSE-OPEN operation to the heat map after thresholding. The opening operation involves erosion followed by dilation and is mainly used to remove isolated small dots and bridges between different regions in the heat map. The closing operation involves dilation followed by erosion and is mainly used to fill in a few concave regions in the patch area that were filtered out by the threshold. The kernel size for the opening and closing operations is adaptively selected based on the image size, more details can be found in the supplementary material.

\section{Defense Evaluation on Digital Attacks}
\label{sec:formatting}

\begin{table*}[t]
\centering
\caption{mAP(\%) under different adversarial patch attacks. The best performance is \textbf{bolded}, and the suboptimal performance is \underline{underlined}.}
\label{tab:overall}
\resizebox{\textwidth}{!}{%
\begin{tabular}{c|c|c|cc|ccc|cccccc}
\hline
\multirow{2}{*}{Detector}                                               & \multirow{2}{*}{Defense} & \multirow{2}{*}{Clean} & \multicolumn{2}{c|}{Localized Noise \cite{liu2018dpatch}} & \multicolumn{3}{c|}{Printable Patch \cite{thys2019fooling}}             & \multicolumn{6}{c}{Natural-looking Patch \cite{hu2021naturalistic}}                                                           \\ \cline{4-14} 
                                                                        &                          &                        & \begin{tabular}[c]{@{}c@{}}DPatch\\ 75$\times$75\end{tabular} & \begin{tabular}[c]{@{}c@{}}DPatch\\ 100$\times$100\end{tabular}          & OBJ            & OBJ-CLS        & Upper          & P1             & P2             & P3             & P4             & P5             & P6             \\ \hline
\multirow{6}{*}{\begin{tabular}[c]{@{}c@{}}Faster\\ R-CNN\end{tabular}} & Undefended               & \textit{96.13}         & \textit{52.52}                & \textit{3.84}                & \textit{50.37}          & \textit{67.40}          & \textit{49.99}          & \textit{60.70}          & \textit{74.30}          & \textit{62.80}          & \textit{73.52}          & \textit{72.23}          & \textit{47.66}          \\
                                                                        & LGS (WACV19) \cite{naseer2019local}             & 96.01                  & 95.96                & 96.06               & 75.34          & 80.10          & 79.57          & 61.34          & 73.69          & {\ul 75.06}    & {\ul 79.60}    & {\ul 74.03}    & 58.66          \\
                                                                        & SAC (CVPR22) \cite{liu2022segment}             & \textbf{96.13}         & {\ul 96.23}          & {\ul 96.16}         & {\ul 80.70}    & {\ul 86.20}    & {\ul 81.05}    & {\ul 62.60}    & 74.00          & 62.80          & 78.74          & 72.41          & 48.00          \\
                                                                        & Jedi (CVPR23) \cite{tarchoun2023jedi}            & 95.97                  & 94.15                & 94.20               & 61.40          & 73.10          & 54.45          & 60.70          & {\ul 75.80}    & 64.30          & 70.15          & 68.22          & {\ul 66.12}    \\
                                                                        & ObjectSeeker (SP23) \cite{xiang2023objectseeker}      & 95.96                  & 52.03                & 4.61                & 49.67          & 64.32          & 48.47          & 57.04          & 66.94          & 53.05          & 71.32          & 66.32          & 38.53          \\
                                                                        & \textbf{PAD (Ours)}            & {\ul 96.11}            & \textbf{96.36}       & \textbf{96.26}      & \textbf{84.55} & \textbf{87.80} & \textbf{88.95} & \textbf{68.40} & \textbf{87.81} & \textbf{85.00} & \textbf{87.56} & \textbf{89.21} & \textbf{83.23} \\ \hline
\multirow{6}{*}{YOLOv3}                                                 & Undefended               & \textit{96.42}         & \textit{66.93}                & \textit{64.04}               & \textit{44.07}          & \textit{78.80}          & \textit{62.92}          & \textit{51.48}          & \textit{42.36}          & \textit{64.93}          & \textit{78.67}          & \textit{64.73}          & \textit{66.70}          \\
                                                                        & LGS (WACV19) \cite{naseer2019local}             & 96.03                  & 96.23                & 95.35               & 60.18          & {\ul 84.63}    & {\ul 83.67}    & {\ul 69.27}          & 74.18          & 68.42          & 78.76          & 63.10          & 73.58          \\
                                                                        & SAC (CVPR22) \cite{liu2022segment}             & 96.08                  & {\ul 96.52}          & {\ul 95.95}         & {\ul 79.39}    & 83.46          & 78.96          & 56.01          & 72.93          & 64.80          & {\ul 84.20}    & {\ul 66.31}    & 67.42          \\
                                                                        & Jedi (CVPR23) \cite{tarchoun2023jedi}            & \textbf{96.60}         & 93.64                & 94.78               & 74.18          & 59.76          & 48.63          & 52.17    & {\ul 75.79}    & {\ul 69.28}    & 69.23          & 62.09          & {\ul 71.69}    \\
                                                                        & ObjectSeeker (SP23) \cite{xiang2023objectseeker}      & 95.82                  & 70.78                & 71.17               & 42.31          & 73.27          & 56.18          & 53.78          & 49.59          & 29.28          & 66.34          & 47.63          & 43.89          \\
                                                                        & \textbf{PAD (Ours)}             & {\ul 96.08}            & \textbf{96.63}       & \textbf{96.51}      & \textbf{85.84} & \textbf{91.06} & \textbf{88.56} & \textbf{78.00} & \textbf{87.38} & \textbf{87.46} & \textbf{89.13} & \textbf{87.76} & \textbf{86.13} \\ \hline
\multirow{6}{*}{YOLOv5s}                                                & Undefended               & \textit{95.72}         & \textit{51.07}                & \textit{37.78}               & \textit{28.47}          & \textit{45.22}          & \textit{41.45}          & \textit{35.96}          & \textit{29.67}          & \textit{38.35}          & \textit{38.30}          & \textit{29.69}          & \textit{36.58}          \\
                                                                        & LGS (WACV19) \cite{naseer2019local}             & 96.04                  & 91.56                & 91.37               & 18.19          & 60.86          & 67.61          & 37.87          & 30.32          & 41.40          & {\ul 61.49}    & 39.61          & 48.49          \\
                                                                        & SAC (CVPR22) \cite{liu2022segment}             & 95.72                  & 92.33                & 92.06               & {\ul 74.21}    & {\ul 77.87}    & {\ul 78.24}    &{\ul 40.25}          & 29.77          & 38.46          & 59.03          & 31.03          & 37.43          \\
                                                                        & Jedi (CVPR23) \cite{tarchoun2023jedi}            & \textbf{96.69}         & 87.70                & 90.65               & 42.96          & 46.88          & 48.79          & 38.10 & {\ul 51.59}    & {\ul 54.11}    & 52.22          & {\ul 44.96}    & {\ul 58.84}    \\
                                                                        & ObjectSeeker (SP23) \cite{xiang2023objectseeker}      & 91.61                  & 50.91                & 38.17               & 35.03          & 39.09          & 43.45          & 37.54          & 37.49          & 38.38          & 48.78          & 35.81          & 33.51          \\
                                                                        & \textbf{PAD (Ours)}             & {\ul 96.17}            & \textbf{93.97}       & \textbf{93.03}      & \textbf{84.01} & \textbf{83.62} & \textbf{84.54} & \textbf{42.01}    & \textbf{58.38} & \textbf{69.87} & \textbf{78.97} & \textbf{67.31} & \textbf{61.08} \\ \hline
\end{tabular}%
}
\end{table*}

\subsection{Evaluation Settings}

\noindent\textbf{Target object detectors and dataset.} In our experiments, we use Faster R-CNN \cite{ren2015faster} with a ResNet-50 \cite{he2016deep} backbone, YOLOv2 \cite{redmon2017yolo9000}, YOLOv3 \cite{redmon2018yolov3}, YOLOv5s \cite{YOLOv5} and YOLOv8n \cite{YOLOv8} as our target object detectors. 
All models are pre-trained on MS COCO \cite{lin2014microsoft}. Since most existing adversarial patch attacks that can be used physically are developed for pedestrian detectors \cite{kim2022defending}, we mainly focus on the INRIA Person dataset \cite{dalal2005histograms} which consists of 614 person detection images for training and 288 for testing. Only test images are adopted since there is no training part in PAD. Experiments on other datasets can be found in supplementary material.

\noindent\textbf{Adversarial patch attacks.} To evaluate the defense effectiveness of PAD against different types of patches, we employ 11 distinct patches generated by DPatch \cite{liu2018dpatch}, YOLO adversarial patch \cite{thys2019fooling}, and Naturalistic Patch \cite{hu2021naturalistic}, covering localized noise, printable, and natural-looking patches. DPatch generates a specific-sized patch ($75\times75$ and $100\times100$ in our experiments) located in the upper left corner of each image, using 200 iterations with a learning rate of 0.01. The YOLO adversarial patch (P1-P6) and Naturalistic Patch (OBJ, OBJ-CLS, and Upper) generate multiple patches of varying sizes and positions based on the detectable pedestrians in the image. We also conduct defense experiments against two more attacks \cite{hu2022adversarial}\cite{huang2023t}, the relevant results can be found in the supplementary material.


\noindent\textbf{Implementation details.} Throughout our experiments, we used fixed hyperparameter values for different patch types without any adjustments. We set $r_{mi}$ to 0.5 in Heatmap fusion, which assigns equal weights to Semantic Independence Localization and Spatial Heterogeneity Localization. The value of $p$ in Adaptive thresholding is set to 0.8. The IoA threshold $t_{m}$ for mask matching is set to 0.5.

We compare PAD with four state-of-the-art adversarial patch defenses: LGS \cite{naseer2019local}, SAC \cite{liu2022segment}, Jedi \cite{tarchoun2023jedi}, and ObjectSeeker \cite{xiang2023objectseeker}, corresponding to denoising-based, external segmenter-based, entropy-based and certifiably robust defenses respectively. For LGS, we set the block size to 30, overlap to 5, threshold to 0.1, and smoothing factor to 2.3. For Jedi, due to the reliance on the prior entropy distribution values of the clean dataset and the patch region, using default parameters in code is less effective, we perform parameter tuning for some patches.

\subsection{Overall Defense Performance}

 In object detection tasks, Average Precision (AP) is a widely used evaluation metric that assesses the area under the Precision-Recall Curve, representing the overall performance of a model. Therefore, we utilize mean Average Precision (mAP) at Intersection over Union (IoU) 0.5 to demonstrate the effectiveness of the attacks and defenses. We conduct experiments on different detectors, attacks, and defenses mentioned above and report the results in Table \ref{tab:overall}. Due to space limitations, it only shows results on Faster R-CNN, YOLOv3, and YOLOv5s. Results on YOLOv2 and YOLOv8n can be found in the supplementary material.

The results demonstrate that PAD achieves the best defense performance against various adversarial patch attacks on different detectors. For natural-looking patches (P1-P6) \cite{hu2021naturalistic}, which are more challenging to detect by both humans and machines, the mAP increases by more than 10\% on average (absolute) compared to the suboptimal method. 

From the experimental results, it can be observed that ObjectSeeker \cite{xiang2023objectseeker} performs poorly under these attacks, some even worse than undefended. This is because ObjectSeeker can only defend against hiding attacks, and the assumption does not hold when encountering multiple patches. SAC \cite{liu2022segment} is best at defending against localized noise patches since its segmenter is trained on noise-like patch data. However, its performance significantly drops when facing natural-looking patches, with almost no defense capabilities against some of the patches. 
The performance of Jedi \cite{tarchoun2023jedi} is unstable,  due to the influence of non-patch high-entropy regions. In contrast, PAD demonstrates robustness against various patches, benefiting from the universality of semantic independence and spatial heterogeneity and the complete independence from prior knowledge of attacks.

We also report the mAP of clean samples after defense in Table \ref{tab:overall}. PAD achieves a similarly high clean performance as the vanilla object detectors (0.02\% drop on Faster R-CNN, 0.34\% drop on YOLOv3, and 0.45\% rise on YOLOv5s). For clean images without patches, although areas with relatively high values may remain after heat map threshold processing, they are usually scattered and will be eroded during subsequent morphological operations, thus not significantly impacting the model's performance.

\begin{figure*}[t]
    \centering
    \includegraphics[width=1\textwidth]{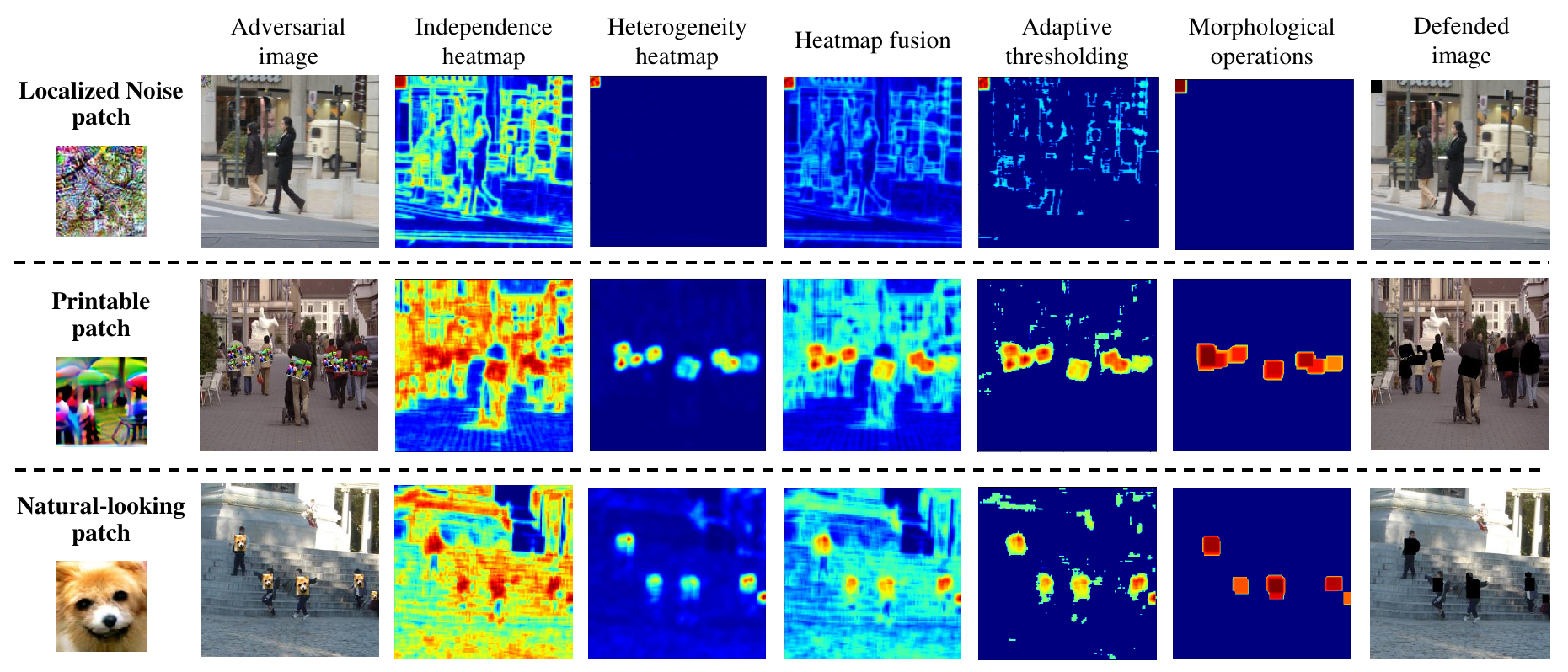}
    \caption{Visualization examples illustrating the patch localization process of PAD across different adversarial patch types.}
    \label{fig:visualization}
\end{figure*}

\subsection{Patch Localization Performance}

Patch localization is a crucial step in the defense process, as it forms the foundation for subsequent patch removal. Therefore, we conducted a further evaluation of the patch localization performance. For accurate quantification, we propose a new metric called Patch Localization Recall. For each ground truth mask $M_{patch}$ corresponding to an introduced patch region, we calculate the IoA between $M_{patch}$ and the generated masks $M_{defense}$ obtained by the defense method, and mark this patch using the following notation:
\begin{equation}
F_{m}=\left\{
	\begin{aligned}
	1 \quad IoA(M_{patch}, M_{defense}) \ge 0.5\\
	0 \quad IoA(M_{patch}, M_{defense}) < 0.5\\
	\end{aligned}
	\right
	.
\end{equation}

\begin{equation}
Recall_{patch} = \frac{\sum_{m=1}^{M} F_{m}}{M}. 
\end{equation}

\begin{table}[t]
\centering
\caption{Patch Localization Recall(\%) on Faster R-CNN.}
\label{tab:plrecall}
\resizebox{0.48\textwidth}{!}{%
\begin{tabular}{cc|ccc}
\hline
\multicolumn{2}{c|}{\diagbox{Attack}{Defense}}                                                      & SAC \cite{liu2022segment}    & Jedi \cite{tarchoun2023jedi}           &  \textbf{PAD}    \\ \hline
\multicolumn{1}{c|}{\multirow{2}{*}{\begin{tabular}[c]{@{}c@{}}Localized\\ Noise \cite{liu2018dpatch}\end{tabular}}}       & DPatch-75  & 100.00 & 9.38           & \textbf{100.00} \\
\multicolumn{1}{c|}{}                                                                                 & DPatch-100 & 100.00 & 40.63          & \textbf{100.00} \\ \hline
\multicolumn{1}{c|}{\multirow{3}{*}{\begin{tabular}[c]{@{}c@{}}Printable\\ \cite{thys2019fooling}\end{tabular}}}          & OBJ        & 33.69  & 28.89          & \textbf{85.90}  \\
\multicolumn{1}{c|}{}                                                                                 & OBJ-CLS    & 38.08  & 34.63          & \textbf{86.24}  \\
\multicolumn{1}{c|}{}                                                                                 & Upper      & 35.68  & 33.95          & \textbf{64.51}  \\ \hline
\multicolumn{1}{c|}{\multirow{6}{*}{\begin{tabular}[c]{@{}c@{}}Natural-\\ looking\\ \cite{hu2021naturalistic}\end{tabular}}} & P1         & 1.33   & 27.30 & \textbf{31.16}           \\
\multicolumn{1}{c|}{}                                                                                 & P2         & 0.00   & 33.16          & \textbf{66.97}  \\
\multicolumn{1}{c|}{}                                                                                 & P3         & 0.93   & 35.68          & \textbf{70.43}  \\
\multicolumn{1}{c|}{}                                                                                 & P4         & 29.69  & 34.75          & \textbf{81.49}  \\
\multicolumn{1}{c|}{}                                                                                 & P5         & 0.67   & 34.62          & \textbf{74.57}  \\
\multicolumn{1}{c|}{}                                                                                 & P6         & 0.93   & 33.16          & \textbf{70.84}  \\ \hline
\end{tabular}%
}
\vspace{-1mm}
\end{table}

Since there is no patch localization process in certifiably robust defense,
and the localization results of LGS \cite{naseer2019local} are not continuous regions, we primarily compare PAD with SAC \cite{liu2022segment} and Jedi \cite{tarchoun2023jedi}. The results are presented in Table \ref{tab:plrecall}, showing a significant improvement (~30\%-55\% absolute and ~2-3x relative) over existing state-of-the-art works. In the case of YOLO adversarial patch \cite{thys2019fooling} and Naturalistic Patch \cite{hu2021naturalistic}, the ground truth masks include small regions that can be easily mistaken for the background. This is because the attack involves covering almost every visible pedestrian in the image with a patch, including small individuals in the distance. As a result, achieving high Patch Localization Recall values becomes more challenging. However, PAD still demonstrates effective performance against these attacks.

SAC \cite{liu2022segment} exhibits good performance for DPatch \cite{liu2018dpatch}, as its segmenter is trained on adversarial images generated with PGD \cite{madry2017towards}, which falls under the category of localized noise. However, it struggles when faced with natural-looking patches that it has not encountered before, leading to almost zero Patch Localization Recall. Jedi \cite{tarchoun2023jedi} performs poorly on DPatch-75, which may caused by the difficulty in the prior entropy distribution values adjustment. In contrast, PAD achieves high Patch Localization Recall for different types of patches, as it does not rely on prior knowledge or existing attack data.

According to the definition of Patch Localization Recall, It is natural to think that a defense method can achieve a high Patch Localization Recall by generating a mask with the widest possible coverage. However, removing a large number of non-patch areas from the image will inevitably result in a decrease in detection mAP. PAD achieves the highest values in both Patch Localization Recall and detection mAP, showcasing superior defense performance. We provide visualization of the patch localization process in Figure \ref{fig:visualization}.

\subsection{Ablation Study}
\label{sec:digital-ablation}

To investigate the individual impacts of semantic independence and spatial heterogeneity in PAD, we conduct an ablation study
that involves using only the $H_{mi}$ from Semantic Independence Localization and only the $H_{cd}$ from Spatial Heterogeneity Localization. Partial results on YOLOv8n are presented in Table \ref{tab:ablation}. It can be observed that the full method, which combines semantic independence and spatial heterogeneity, achieves more stable overall performance.

\begin{table}[h]
\centering
\caption{mAP (\%) of ablated defenses on YOLOv8n.}
\label{tab:ablation}
\resizebox{0.48\textwidth}{!}{%
\begin{tabular}{l|ccccc}
\hline
\multicolumn{1}{c|}{\diagbox{Defense}{Attack}} & OBJ           & Upper         & P3            & P4            & P5            \\ \hline
LGS                                 & 47.5          & 82.0          & 53.1          & 79.4          & 62.4          \\
SAC                                 & 81.9          & 58.1          & 51.8          & 78.2          & 53.5          \\
Jedi                                & 57.6          & 84.6          & 66.9          & 65.9          & 64.2          \\
\textbf{PAD}-MI only                        & 78.6          & 86.7          & {\ul 76.3}    & 77.4          & 74.0          \\
\textbf{PAD}-CD only                        & {\ul 86.3}    & \textbf{89.8} & 76.1          & {\ul 85.4}    & \textbf{82.2} \\
\textbf{PAD}-all                            & \textbf{87.5} & {\ul 88.9}    & \textbf{78.7} & \textbf{85.4} & {\ul 81.5}    \\ \hline
\end{tabular}%
}
\vspace{-3.5mm}
\end{table}

Additionally, we have observed that spatial heterogeneity tends to outperform semantic independence in digital attack experiments. In digital attacks, patch generation is completely independent of the original clean image and less affected by interference, resulting in more pronounced heterogeneity, leading to better performance. However, in physical attacks where the patch is physically printed and imaged alongside other parts of the scene, the manifestation of heterogeneity may become weaker, and the role of semantic independence becomes more significant. In such cases, semantic independence outperforms spatial heterogeneity.

In this paper, to validate the defense performance of PAD without any parameter tuning, equal weights are assigned to Semantic Independence Localization and Spatial Heterogeneity Localization. By adjusting the weight allocation for digital attacks and physical attacks respectively, PAD can achieve even better results.

\vspace{-0.4em}
\section{Defense Evaluation on Physical Attacks}
\label{sec:formatting}

To validate the effectiveness of PAD against physical attacks, we conducted experiments using a publicly available dataset that consists of physical adversarial patches \cite{braunegg2020apricot}, as well as physical attack videos captured by our own. We kept all parameters unchanged, ensuring that the implementation details remained consistent with the digital experiments.

\subsection{Evaluation on APRICOT}

APRICOT \cite{braunegg2020apricot} consists of 1,011 photos with high resolution captured in real-world environments, encompassing both indoor and outdoor scenes. Each photo contains a printed physical adversarial patch, which varies in size, shape, location, viewing angle, and lighting conditions. These patches are generated to cause false detection of non-existent objects, targeting 10 specific classes.

\begin{figure}[t]
    \centering
    \includegraphics[width=0.48\textwidth]{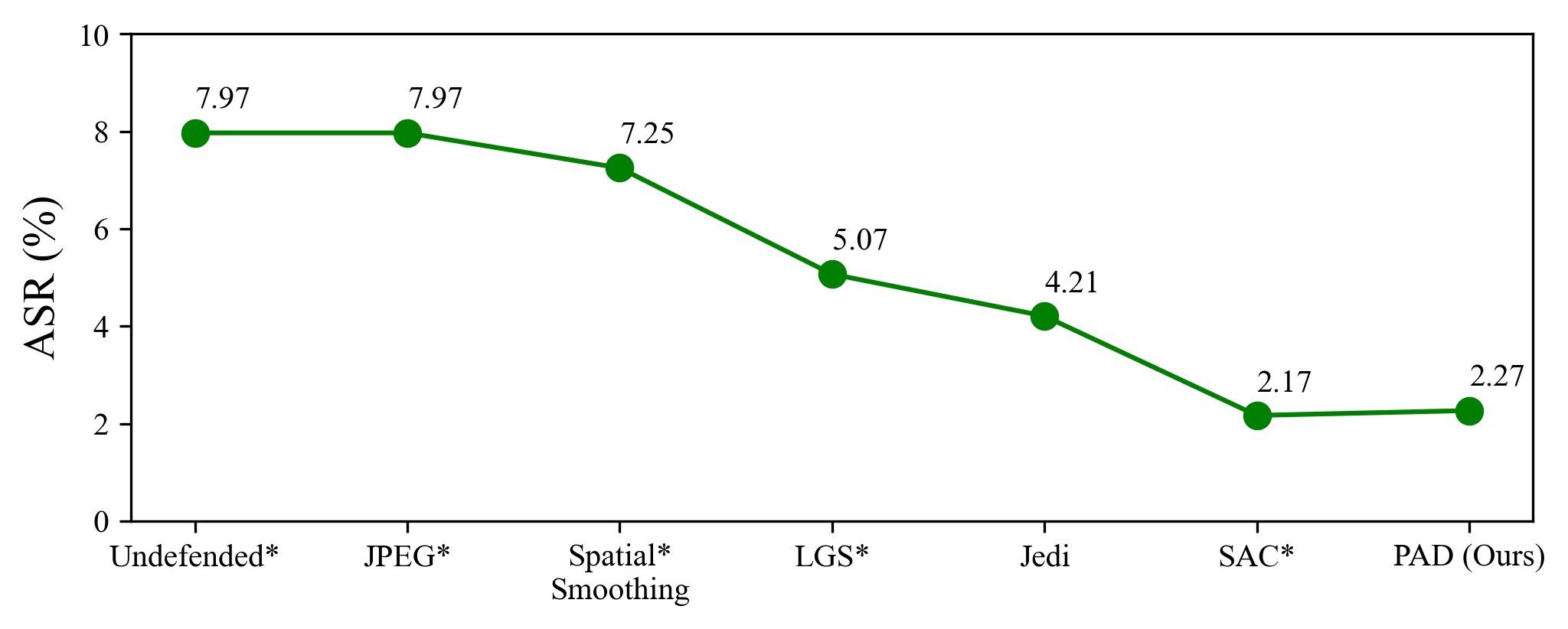}
    \caption{ASR (\%) after defenses, lower values indicate better defense performance. Results with * are from \cite{liu2022segment}.}
    \label{fig:apricot-asr}
\end{figure}

We use Faster R-CNN \cite{ren2015faster} model pretrained on MS COCO \cite{lin2014microsoft} as our target object detector and evaluate the defense performance on the development set. Since this is a targeted attack, we use the Attack Success Rate (ASR) as our evaluation metric, setting the IoU threshold to 0.10 and the confidence threshold to 0.30. We present the results after applying different defense methods in Figure \ref{fig:apricot-asr}.

\begin{figure}[t]
    \centering
    \includegraphics[width=0.48\textwidth]{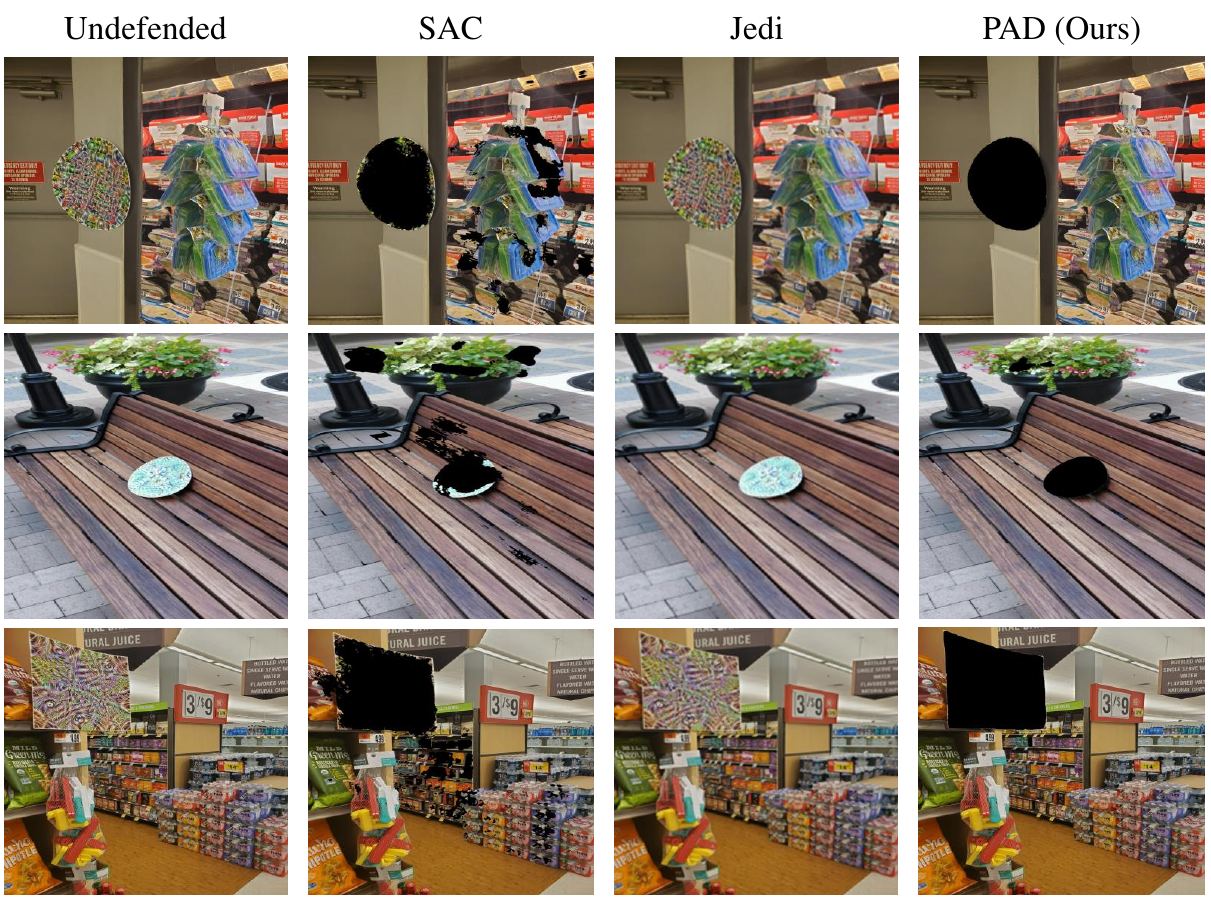}
    \caption{Comparison of defended images on APRICOT. Jedi \cite{tarchoun2023jedi} fails to locate patches, the drop in ASR is caused by resizing in Auto-Encoder. SAC \cite{liu2022segment} masks out most patch regions, but still affects many other regions despite the training with accurate mask annotations. PAD achieves the best removal performance.}
    \label{fig:apricot-result}
\vspace{-1.5mm}
\end{figure}

The results show that PAD, without any prior attack knowledge or training data of adversarial patches, significantly reduces the attack success rate to 2.27\%. Moreover, SAC \cite{liu2022segment} utilizes APRICOT data with accurate masks to train its patch segmenter, getting an attack success rate of only 0.1\% lower than PAD, highlighting the superiority of PAD. We provide examples of defended images produced by different defense methods in Figure \ref{fig:apricot-result}, demonstrating our robustness against physical world patches of various sizes, shapes, lighting conditions, and angles.


\subsection{Evaluation on physical attack videos}

To further evaluate the effectiveness of PAD against a wider range of patch types in the physical world, we print nine different patches, including P1-P6 \cite{hu2021naturalistic}, OBJ, OBJ-CLS, and Upper \cite{thys2019fooling}, and capture videos in five different indoor and outdoor scenes while holding these patches.

Due to the significant impact of lighting, distance, and angles on the success rate of physical attacks, we conduct extensive practical filming and testing to select a subset comprising images with relatively higher attack success rates. The final defense test set consists of 1100 photos, more details about the data distribution can be found in the supplementary material.

\begin{figure}[t]
	\centering
	\begin{subfigure}{0.50\linewidth}
		\centering
		\includegraphics[width=0.95\linewidth]{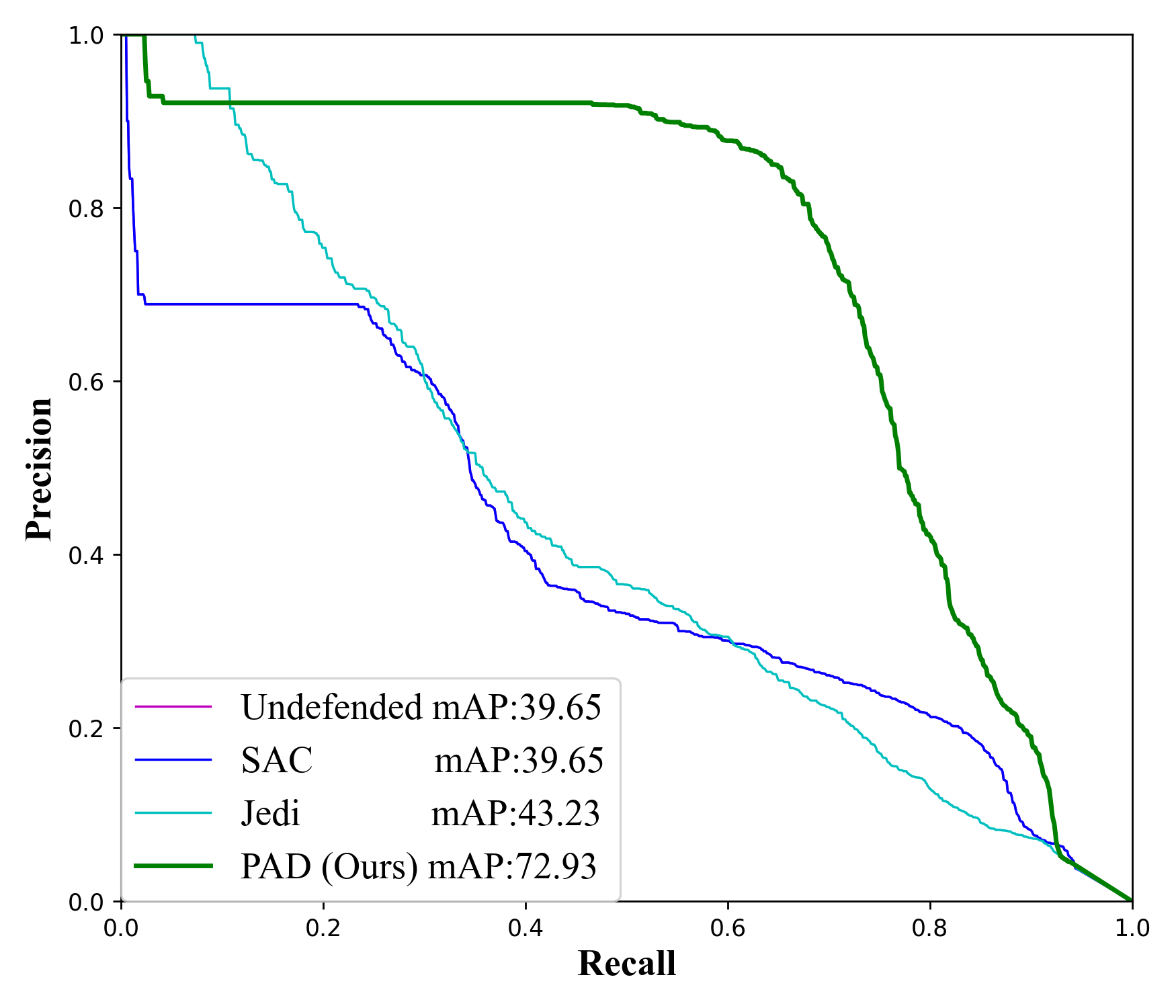}
		\caption{PR-curve of different defenses. SAC coincides with Undefended.}
		\label{physical-prcurve}
	\end{subfigure}
	\centering
	\begin{subfigure}{0.49\linewidth}
		\centering
		\includegraphics[width=0.95\linewidth]{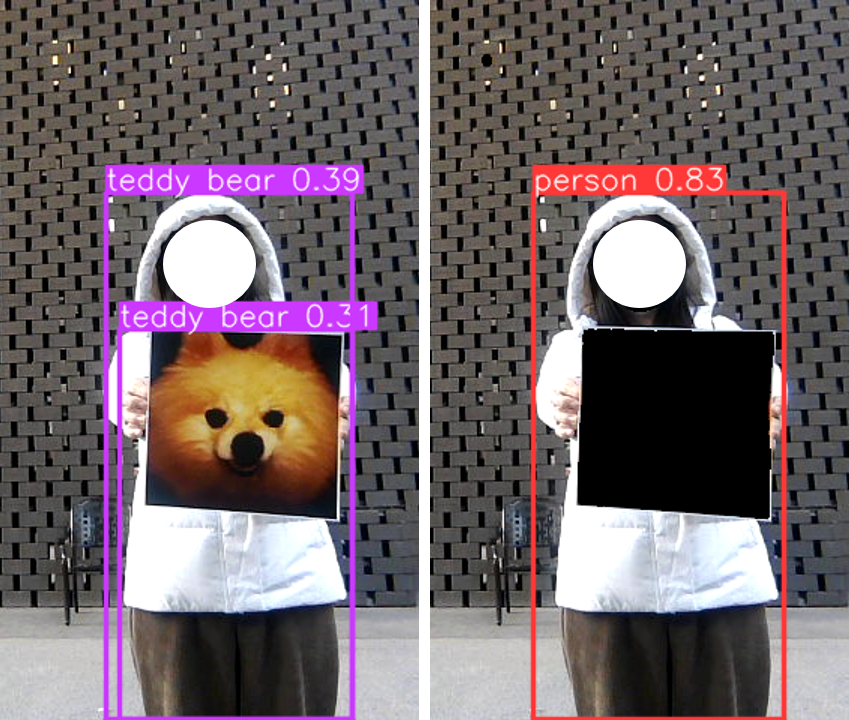}
		\caption{Original and defended image. White circles added after prediction.}
		\label{physical-image}
	\end{subfigure}
  \caption{Defense results on our physical test set. }
  \label{fig:physical-result}	
\vspace{-2mm}
\end{figure}

We use YOLOv8n as the target object detector and compare the defense performance of PAD with Jedi \cite{tarchoun2023jedi} and SAC \cite{liu2022segment} on this test set. The PR curves in Figure \ref{physical-prcurve} demonstrate the superiority of PAD over the compared state-of-the-art methods. We show an example of the test image and defense result in Figure \ref{physical-image}.

\section{Conclusion}

In this paper, we identify two inherent characteristics of adversarial patches that are independent of their appearance, shape, size, location, and quantity. Leveraging these characteristics, we propose a patch-agnostic defense (PAD) method, which perform adversarial patch localization and removal without prior attack knowledge. PAD offers patch-agnostic defense against a wide range of adversarial patches, significantly enhancing the robustness of various pre-trained object detectors. Without training, PAD eliminates the reliance on existing attack data, making it more adaptable and capable of defending against novel patch attacks that have not been encountered yet. Our experimental results demonstrate the effectiveness in both digital space and the physical world, highlighting the practicality of PAD across different attack scenarios.

\vspace{3mm}
\noindent\textbf{Acknowledgements.} This work is supported in part by the National Natural Science Foundation of China Under Grants No.62176253 and No.U20B2066.

{
    \small
    \bibliographystyle{ieeenat_fullname}
    \bibliography{main}
}


\end{document}